\documentclass[conference]{IEEEtran}
\IEEEoverridecommandlockouts

\usepackage{amssymb}
\usepackage{lineno}

\usepackage{graphics} 
\usepackage{epsfig} 
\usepackage{mathptmx} 
\usepackage{times} 
\usepackage{amsmath} 
\usepackage{amssymb}  
\usepackage{multirow}
\usepackage{multicol}
\usepackage{hyperref}
\usepackage{makecell}
\usepackage{textgreek}
\usepackage{forest}
\usepackage{lipsum}

\usepackage{float}
\restylefloat{table}

\begin{document}

\title{
    Atlas Fusion - Modern Framework for \\ Autonomous Agent Sensor Data Fusion
    \thanks{
        The work has been performed in the project NewControl: Integrated, Fail-Operational, Cognitive Perception, Planning and Control Systems for Highly Automated Vehicles, under grant agreement No 826653/8A19006. The work was co-funded by grants of Ministry of Education, Youth and Sports of the Czech Republic and  Electronic Component Systems for European Leadership Joint Undertaking (ECSEL JU)
        The work was supported by the infrastructure of RICAIP that has received funding from the European Union's Horizon 2020 research and innovation programme under grant agreement No 857306 and from Ministry of Education, Youth and Sports under OP RDE grant agreement No CZ.02.1.01/0.0/0.0/17\_043/0010085.
        
        Copyright notice: 978-1-6654-6726-1\/22\$31.00 \copyright2022 IEEE
    }
}

\author{
    \IEEEauthorblockN{1\textsuperscript{st} Adam Ligocki}
    \IEEEauthorblockA{\textit{Cybernetics and Robotics, CEITEC} \\
    \textit{Brno University of Technology}\\
    Brno, Czechia \\
    0000-0002-6813-4318}
    \and
    \IEEEauthorblockN{2\textsuperscript{nd} Aleš Jelínek}
    \IEEEauthorblockA{\textit{Cybernetics and Robotics, CEITEC} \\
    \textit{Brno University of Technology}\\
    Brno, Czechia \\
    0000-0001-7519-2092}
    \and
    \IEEEauthorblockN{3\textsuperscript{rd} Luděk Žalud}
    \IEEEauthorblockA{\textit{Cybernetics and Robotics, CEITEC} \\
    \textit{Brno University of Technology}\\
    Brno, Czechia \\
    0000-0001-5996-8137}
}

\maketitle

\begin{abstract}
    In this paper, we present our software sensor fusion framework for self-driving cars and other autonomous robots. We have designed our framework as a universal and scalable platform for building up a robust 3D model of the agent's surrounding environment by fusing a wide range of various sensors into the data model that we can use as a basement for the decision making and planning algorithms. Our software currently covers the data fusion of the RGB and thermal cameras, 3D LiDARs, 3D IMU, and a GNSS positioning. The framework covers a complete pipeline from data loading, filtering, preprocessing, environment model construction, visualization, and data storage.  The architecture allows the community to modify the existing setup or to extend our solution with new ideas. The entire software is fully compatible with ROS (Robotic Operation System), which allows the framework to cooperate with other ROS-based software. The source codes are fully available as an open-source under the MIT license. See \url{https://github.com/Robotics-BUT/Atlas-Fusion}.
\end{abstract}


\section{Introduction}

As the world is diving deeper into the problem of self-driving cars and other autonomous robots, there is a large number of sophisticated systems for analyzing data and controlling the specific problems of autonomous behavior. However, these systems, like \cite{kato2018autoware} or \cite{li2019aads} are very complex and require dozens of hours to understand the architecture and to be able to start to develop a new solution on top of the existing one. 

As members of the academic community, we are experimenting with many new approaches. Our primary motivation is to search for new ways and improve the current state-of-the-art techniques. For this purpose, we designed a system aiming at surrounding environment sensing and map building in mobile robotics.

As a result, our team created this C++ framework focusing on data fusion from the various sensor types into a robust representation of the robot's surroundings model.

\begin{figure}[t]
    \centering
    \includegraphics[width=8.8cm]{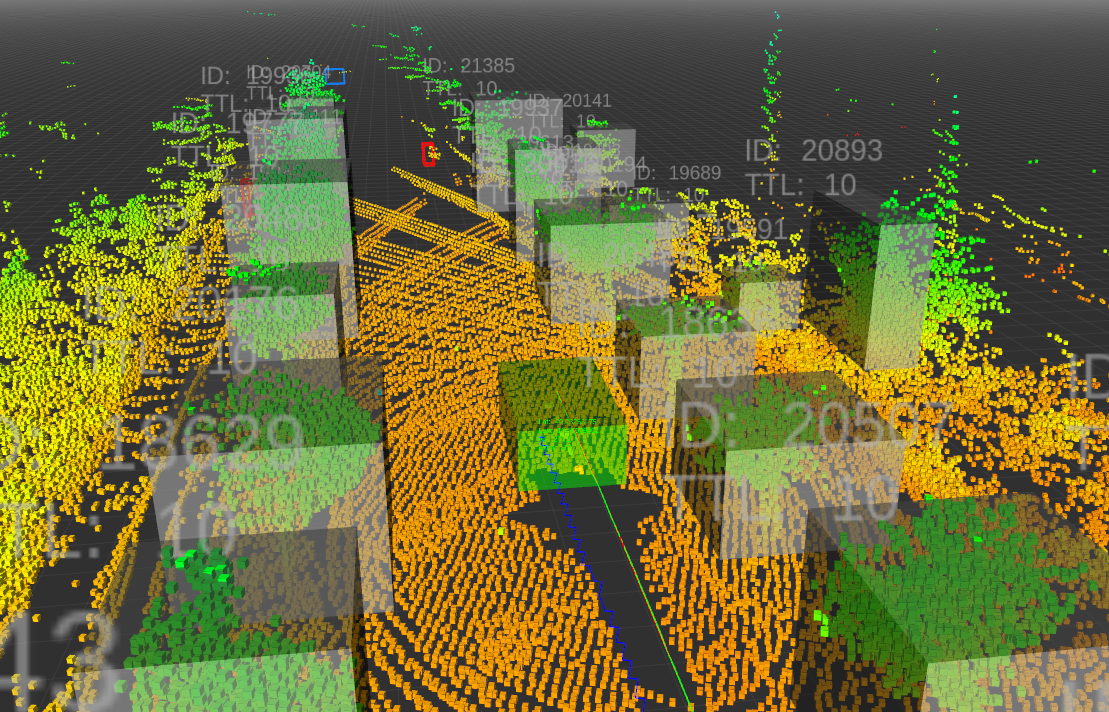}
    \caption{The example of the RViz visualization of the runtime model of the surrounding environment. Grey boxes are the LiDAR-based detections, and color frustums are the RGB images' neural network detections. The green object at the center represents the agent and the lines behind the agent are the trajectories estimated by different filtering algorithms.}
    \label{fig:overview}
\end{figure}

\section{General Architecture Description}

We have designed the software with the idea of a very minimalistic pipeline and simple modification to develop and deploy new ideas and algorithms quickly. 


\subsection{Input Data}

As an input data format, we have chosen the same representation used previously in our work on Brno Urban Dataset \cite{ligocki2019brno}, which is inspired by \cite{maddern20171}. 

The data are stored as an h265 video in case of RGB and thermal camera data, .ply files for LiDAR scans, and CSV data files for GNSS, IMU, and camera and LiDAR timestamps. 

\subsection{Core Pipeline}

At the startup, the program reads the basic configuration from the config file. The configuration provides a path to the offline record, and the data loading module loads up all the necessary information for offline data interpreting. After that, the main pipeline begins.

The loading module loads all timestamped data into the memory and later provides the data in the correct order, one by one. The pipeline redirects data into the dedicated processing section based on the data type and from which sensor the data comes. The output data, like detected obstacles, static obstacles, or moving entities, are stored in the local map data model.

The entire pipeline has a linear architecture, so the data processing algorithms are sorted one by one. This waterfall-like design allows anybody to add or remove a new data processing algorithm without affecting the current ones.

\subsection{Outputs}
The framework's main output is the map of the surroundings, stored in the Local Map block, with the precise detection of the possible static and dynamic obstacles. The following decision-making algorithms can use this map to adjust the agent's behavior based on the mapping process's data. 

Secondary, there are several other outputs described in detail in section~\ref{sec:data_processing_pipelines}. There are the 3D models of all the places that the agent visited during the mapping session, projections of neural network's detection from RGB cameras to thermal images (an annotated IR dataset for object detection is created this way), and the depth maps for camera images generated from the aggregated point clouds. 

\section{Modules}

\subsection{Data Loaders} \label{subsec:data_loaders}
As our framework is currently not working with online data, there is an interface that loads stored records and provides the loaded data ordered by their timestamps to the main pipeline.

There is a data loader for every physical sensor that reads only one data series. These data loaders are wrapped by a central data loader that creates an interface between stored data and the main pipeline. All the data loaders have ordered the timeline of their data series. When the main pipeline is ready to accept the next data packet, the central data loader asks all the loaders for their smallest timestamp. The data loader with the lowest timestamp will provide the data packet to the processing pipeline. 


\subsection{Data Models}

The first part of the data models is the raw input data representation. Every sensor has one or more classes that cover the range of the input data. For example, a camera. There are two classes \url{CameraFrameDataModel} for RGB image representation and the \url{CameraIRFrameDataModel} for the thermal camera image data entity. Every instance of those classes defines the camera sensor identifier, precise timestamp, image frame, and optionally pre-generated YOLO neural network object detections. This data packet keeps all the important information, and the data loader passes the instance of this class when the main processing pipeline requests the latest image data.

The second part of the data models is the internal data representation models used for communication between the modules in the primary data processing pipeline. For example, the \url{Lidar Detection} structure for objects detected in the LiDAR domain, \url{Local Position} as a relative metric position w.r.t. the origin of the mapping session, \url{Frustum Detection} for the camera-based detected objects and many others.

\subsection{Algorithms}

The "Algorithms" module is the core one. It contains all the data processing code. Here the implemented classes cover the agent's position filtration based on Kalman fusion of the GNSS and IMU inputs, functionality for projecting objects from the 3D environment into the camera frames and back, generating a depth map from the LiDAR data, or the redundant data filtration. The "Algorithms" module is the main section where the implementation of the pipelines is described in Section~\ref{sec:data_processing_pipelines}.

\subsection{Local Map}

The "Local Map" module primarily represents the software that holds the internal map of the surrounding environment. There are two main classes. The first one is \url{Local Map}. This class is a simple container that allows us to store and read out data models of the map representation entities, like aggregated LiDAR model of the near surrounding, detected obstacles, YOLO detections, and higher representations of the more complex fused data. The second class is \url{Objects Aggregator}. This class fuses low complexity detections, such as LiDAR and camera-based detected objects, into the higher complexity representation, fusing geometrical shape information, object type, kinematic model, motion history, etc.


\subsection{Visualizers}

This module handles the interface between the main pipeline and local map, and the rendering engine. The main class, called \url{VisualizationHandler} provides a wrapper over the entire rendering logic. For every specific data type (IMU data - \url{ImuVisualizer}, camera frames - \url{CameraVisualizer}, point clouds - \url{LidarVisualizer}, etc.) there is dedicated class that manages the interface between the central point and the visualization engine (RViz in our case). 


\subsection{Data Writers}

The Data Writer section covers the classes responsible for writing Local Map data to the local hard drive storage. Currently, there are the implementations for saving the aggregated LiDAR point cloud projected to the camera plain (see \ref{subsec:agg_lidar_to_image}) and the class for storing RGB YOLO detections projected into the thermal camera (see \ref{subsec:rgb_yolo_to_ir}).
\section{Data Processing Pipelines} 
\label{sec:data_processing_pipelines}

The framework implements several principles of data processing and map building. In this section, we are describing the basics of the most important ones.

\subsection{Precise Positioning}

Without precise positioning, it would be impossible to build a reliable map model and aggregate information in time.

For our purpose, we used the differential RTK GNSS that samples a global position with the precision of one $\sigma$ below 2cm and provides an azimuth of the measurement setup. To improve the dynamic positioning, we also use the linear acceleration and angular velocity from the IMU sensor. An example of the fusion of these sensors could be \cite{caron2006gps}.

The pipeline has the following input data: the global position and heading from the GNSS receiver, linear acceleration, angular velocity, and filtered absolute orientation from the IMU sensor. The IMU automatically compensates for the roll and pitch drift by the gravity's direction, and the yaw drift compensates by the magnetic field measurement.

In the beginning, the first GNSS position sets up an anchor that defines the mapping session's origin. This first global position is the origin (the anchor) of the local coordinate system. The core of the position estimation process is the set of 1D Kalman filters \cite{kalman1960new}, \cite{thrun2002probabilistic}, that model position and speed in all three axes of the given environment. Every new incoming GNSS position is converted to the local coordinate system w.r.t. the anchor. This local position is used as a correction for the Kalman filters \cite{terejanu2013discrete} in all three axes. 

\begin{figure}[ht]
    \centering
    \includegraphics[width=8.8cm]{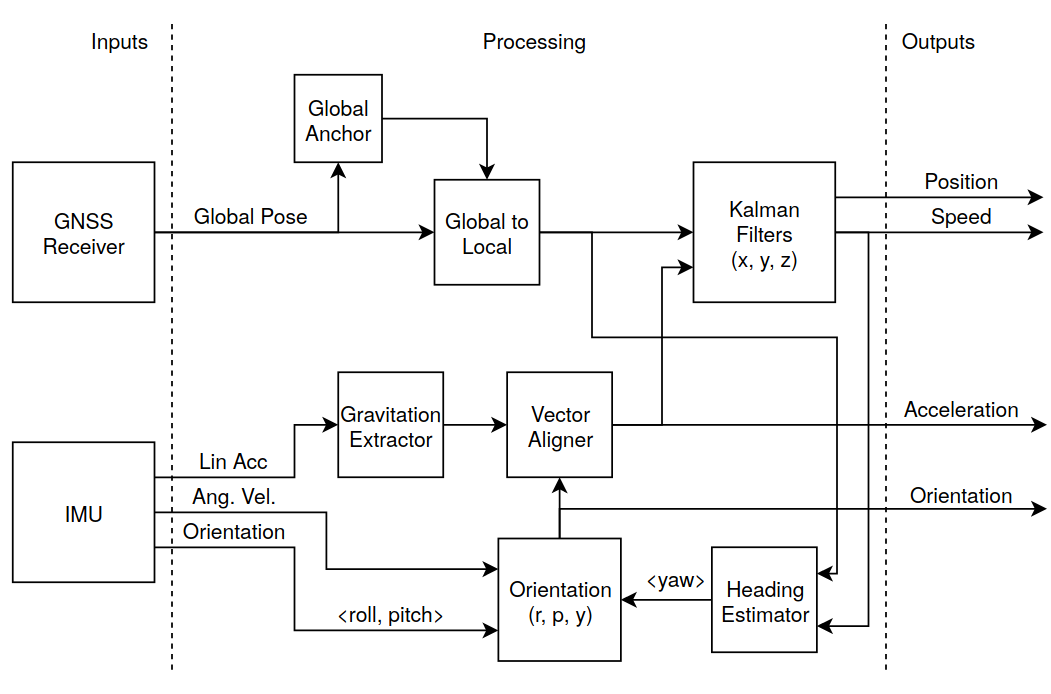}
    \caption{Scheme of the position estimation pipeline.}
    \label{fig:positioning_pipeline}
\end{figure}

As the system models the IMU orientation separately on the IMU's internal model, every new angular velocity data system updates its internal model to have a fast response. However, there is always a long-term drift for this long-term noisy data integration. The system fuses its internal model with the IMU's one using the low pass filter to remove the roll and pitch drifts. To compensate for the yaw drift, it combines the heading measured by the GNSS receiver and its differential antennas with the heading estimated by the agent's speed, which the motion model estimates.

\subsection{LiDAR data aggregation} \label{subsec:lidar_data_agg}

As we are using the rotating 3D LiDARs, the scanners perform measurements in different directions at different times during the scanner motion, and the robot is constantly changing its position. All these effects cause the outcome measurement to be significantly distorted \cite{merriaux2017lidar}, \cite{zhang2019point}.  

Thus, we can not merge all the scans into one because the result would be inaccurate and blurred.

The input LiDAR data could come from several LiDAR scanners. The entire process assumes that each scan stores the data in the same order as it was measured. 
However, the input data are at the beginning filtered by the data model's callback and downsampled by the \url{Point Cloud Processor} call instance to reduce the computational complexity of the later point cloud transformation. At the same time, the positioning system provides the agent's position when the current and the previous scans were taken.


All these three information, the scan, and both positions are passed to the \url{Point Cloud Extrapolator} instance. There the point cloud is split linearly into the N batches of the same size. Because the scan data are sorted, each batch covers a small angular section of the entire scan, corresponding to the small-time period when the batch data has been taken.

\begin{figure}[ht]
    \centering
    \includegraphics[width=8.8cm]{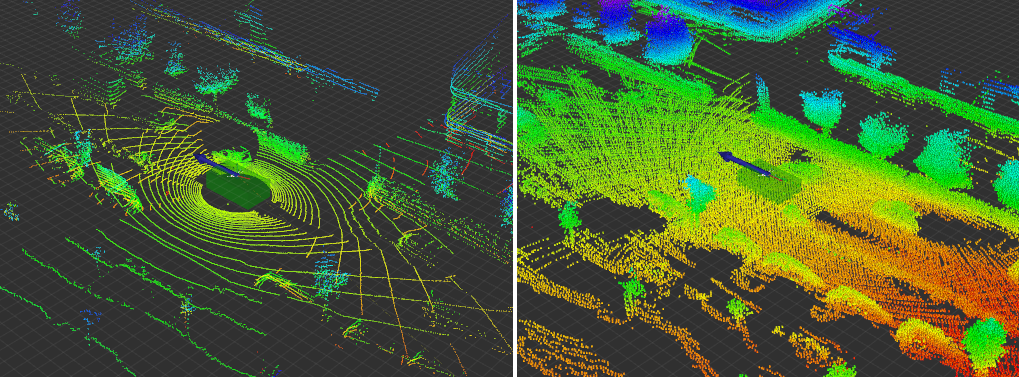}
    \caption{Comparison of the non-aggregated point cloud from two Velodyne HDL-32e scanners (left) and the aggregated ones (right) on the aggregation period of 1.5s. }
    \label{fig:pc_agg_nonagg}
\end{figure}

We have already estimated the valid transformation for every batch for a short time when the batch's data has been scanned. This transformation corresponds to the IMU position w.r.t. the origin of the local coordinate system. Thus, we have to aggregate one more transformation that expresses the frame difference between the given LiDAR sensor and the IMU reference frame. In this way, we can calculate the final homogeneous transformation transform every single point cloud measurement from the scanner's frame to the local coordinates frame. However, transforming every single point is very demanding on computational power. The points are not transformed immediately, but the batch holds the data in the original frame, and the transformation could be evaluated later in the pipeline.

\subsection{Camera-LiDAR Object Detection}

LiDAR can measure the distance and the geometrical shape of the obstacle with high accuracy. On the other hand, to be able to recognize the specific class of the object based only on the point cloud and geometrical shapes is quite challenging. The very opposite of this approach is object detection on the camera images. These days, neural networks can localize and classify objects on the RGB images in real-time with several dozens of fps \cite{bochkovskiy2020yolov4}. However, although we have quite a reliable object classification and localization in the 2D plane, it is tough to estimate the detected object's distance.  

\begin{figure}[ht]
    \centering
    \includegraphics[width=8.8cm]{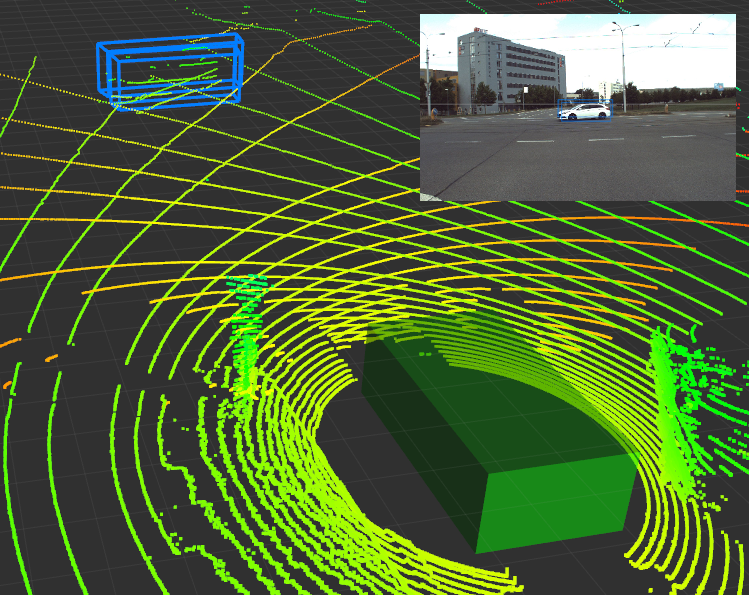}
    \caption{Car detected by the neural network in both frontal cameras. Distance of the 2D detection is estimated based on the aggregated LiDAR data. Camera view in the right top corner.}
    \label{fig:frustum_detection}
\end{figure}

For this purpose, we have created a system that fuses the LiDAR data and camera detections and combines them into a single representation. 

There is an estimated median distance of the LiDAR measurements projected to the detection bounding box for every detection on the RGB image. This information system generated the 3D frustum representation in the output map of the detected obstacle.

\subsection{RGB YOLO Detections to IR Image} \label{subsec:rgb_yolo_to_ir}

If we focus on the field of neural network training, we can find a large number of papers \cite{redmon2018yolov3}, \cite{liu2016ssd}, \cite{zhao2019object} that deal with object detection on RGB images. However, not many works focus on thermal images \cite{agrawal2019enhancing}. The thermal domain is very beneficial for autonomous agents because it allows agents to sense their surroundings even in bad lights or weather conditions.

There is not only a smaller number of works interested in the learning neural networks to detect objects on the thermal images  \cite{ivavsic2019human}, \cite{herrmann2018cnn} compared to the visible light spectrum, but also the there is also a dramatically smaller background in existing datasets. There are very few publicly available sources of annotated thermal images that could be used for training purposes, like KAITS \cite{jeong2019complex} or the FLIR \cite{flir}.

We have proposed a system that would automatically generate a large amount of annotated IR images based on the transferring object detections from the RGB images to the thermal ones, which will help in the future when we will train neural networks for in the thermal image domain \cite{pratt1993discriminability}. 

\begin{figure}[ht]
    \centering
    \includegraphics[width=8.8cm]{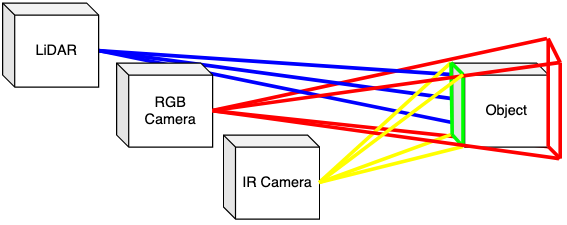}
    \caption{1 (red) - the YOLO neural network detects objects in the RGB image. This 2D detection can be represented as a 3D frustum in the real world. 2 (blue) - the LiDAR measures object distance. 3 (green) - by combining LiDAR data and 3D frustum, we can estimate the frontal plane of the detected object. 4 (yellow) - the detected object's plane is reprojected into the IR camera.}
    \label{fig:yolo_rgb_to_ir}
\end{figure}

The basic idea is to preprocess the detections on the RGB camera, which is semantically close to the IR camera and oriented in the same direction. The nearest IR frame in time is taken for every RGB frame for which the object detection has been performed. In the next phase, the aggregated point cloud model (see \ref{subsec:agg_lidar_to_image}) is used to estimate the distance of the detected obstacle so that the obstacle can be transformed from the 2D image plane into the 3D model of the environment. The last phase is to project the 3D modeled obstacle's frontal face into the thermal image and store the parameters of the projected objects in the same format as the origin RGB detections do.


\subsection{Aggregated LiDAR Data to Image Projection} \label{subsec:agg_lidar_to_image}

Currently, many academical publications deal with convolutional neural networks and improve the performance of those state-of-the-art algorithms. However, there is a large number of papers that cover the RGB image object detection, but much less of those that would be dealing with the object classification and detection in the IR (thermal) domain \cite{rodin2018object} and even less that would try to process the depth images \cite{ophoff2019exploring}.

Our framework allows us to merge all these three domains into a single one.


Every new frame from the thermal camera triggers the following process. From the motion model, the current position of the IMU in the local coordinate system is requested. Simultaneously, the transformation between the IMU and the IR camera is known from the calibration frame.

\begin{figure}[ht]
    \centering
    \includegraphics[width=8.8cm]{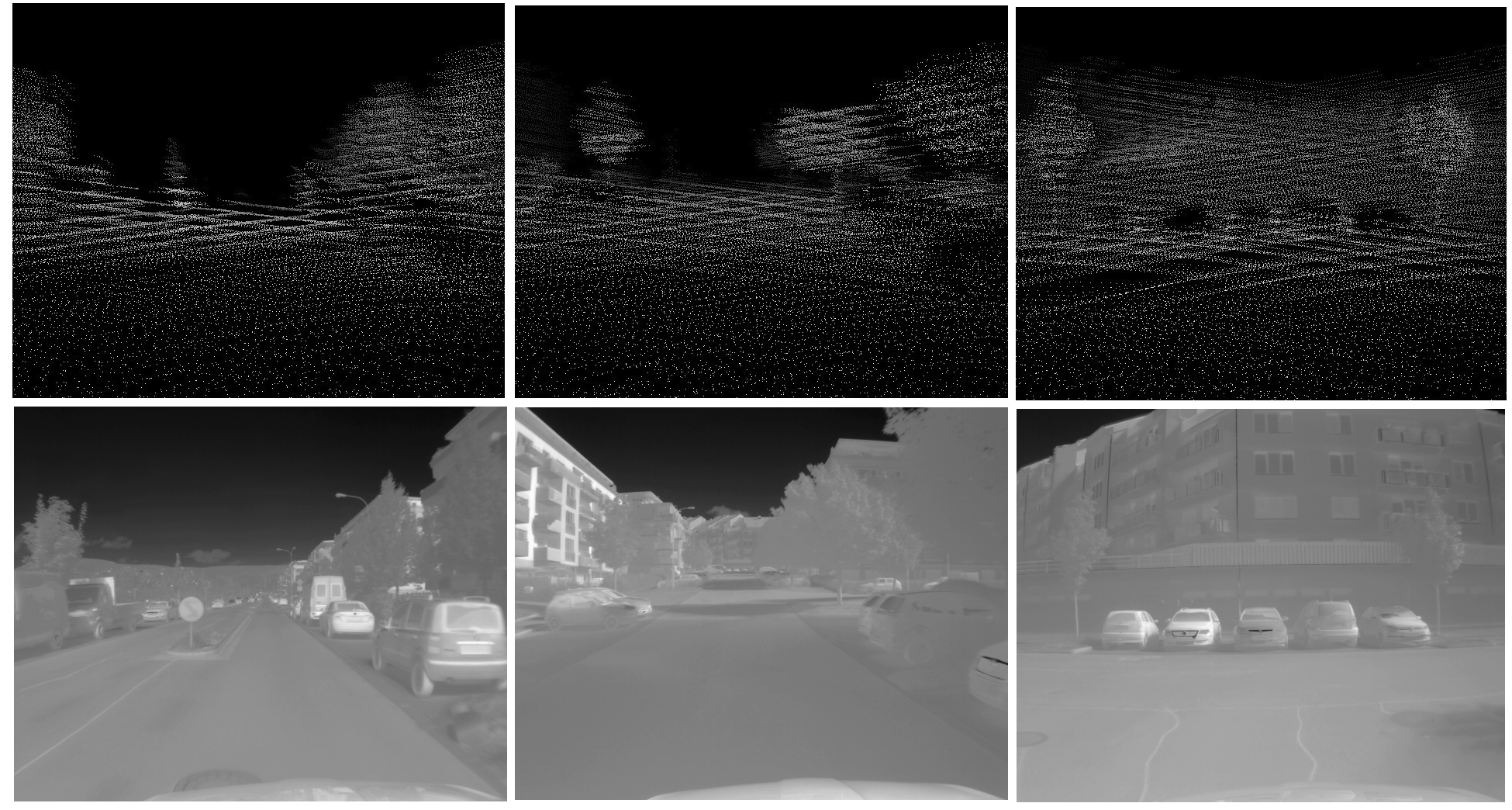}
    \caption{Example of depth images generated based on the aggregated point cloud model. Depth images (top) paired with the corresponding thermal images (bottom). Point cloud has been projected to the camera frame. The same technique can be applied also on RGB images.}
    \label{fig:depth_ir_overview}
\end{figure}

From the \url{Point Cloud Aggregator}, the currently aggregated set of the point cloud batches is requested and passed into the instance of the \url{Depth Map} class. The \url{Depth Map} is also provided by the current position and the IMU to camera transformation and the camera calibration parameters. By combining all this information, for every point cloud batch, there is applied additional transformation. The entire transformation chain is currently following from the LiDAR frame to the IMU frame to the Origin frame to the IMU frame to the IR Camera frame. 

\subsection{Visualizations}

The entire mapping process requires a detailed visualization backend to correctly understand every step of the data processing and the final output environment model. For this purpose, we have used RViz - the visualization tool of the ROS toolkit. It supports elementary geometry objects like points or lines and more complex shapes, like arrows, polylines, and complex visualizations, like point clouds, occupancy grids, or the transformation trees. A single class \url{Visualization Handler} processes the visualization that wraps the entire visualization logic.
\section{External Dependences}

The ROS (Robotic Operating System) \cite{quigley2009ros} makes it possible to communicate with other programs with well-defined API. The entire framework visualization is also realized via the Rviz program, a part of a ROS environment.

For the underlying data representation, like N-dimensional vectors, rotation angles, matrices, quaternions, bounding boxes, frustums, transformations, etc., we have used the previous work of one of the authors, the Robotic Template Library, the C++17 built on the Eigen library. RTL is available at \url{https://github.com/Robotics-BUT/Robotic-Template-Library}. Next to the fundamental data primitives representation, RTL also provides several algorithms for point cloud segmentation and vectorization \cite{10.1007/s11554-016-0562-6}, \cite{10.5220/0005962902160223}

To solve the assignment problem, we have used the implementation \url{https://github.com/aaron-michaux/munkres-algorithm} that refers \cite{Pilgrim1995}.

\section{Future Work}

We have designed our framework in a way that the architecture allows anybody to modify or extend the existing solution. We have put a special effort into building up an abstract system that allows us to scale the current solution to a much larger solution with a reasonable amount of additional complexity. For example, there is no need to modify existing data models and loaders to implement the new sensor's data. We can extend the current software with a few new lines of code based on the given templates. The same we can say about the processing pipelines.

In the future, we are preparing to add more sensors, like radar or ultrasound sensors, extending the current pipeline with the disparity map generation based on the two frontal cameras, optical odometry, or semantic scene segmentation by the neural networks.

We would also like to make this project fully open-source because we believe that these projects can reach a more significant number of developers and researchers, and the bigger community means a more dynamic development process. Our target is to provide a research platform for a large research community that will not need to develop many of those algorithms from scratch and will be able to improve more specific problems for the autonomous robot or the self-driving car domain. 

\section{Conclusion}
As a result of the research project, we have created the experimental mapping framework that allows easy and fast prototyping of new approaches in autonomous agents. The primary data processing pipeline is a single thread with a waterfall-like architecture, making it easy to understand how the data are processed. The modification does not require complicated code refactoring.

The essential parts of our framework are the precise positioning system that fuses GNSS and IMU data. The LiDAR scans aggregator allows us to integrate multiple point clouds into a single dense environment model. Next is the point cloud to camera projection and depth image generating, the point cloud obstacle detection, YOLO neural network-based 3D obstacle detection, and RGB to IR neural network detection mapping.

\bibliography{main.bib}
\bibliographystyle{IEEEtran}

\end{document}